\documentclass[letterpaper]{article} 
\usepackage[arxiv]{aaai23}  
\usepackage{times}  
\usepackage{helvet}  
\usepackage{courier}  
\usepackage[hyphens]{url}  
\usepackage{graphicx} 
\usepackage{natbib}  
\usepackage{caption} 
\DeclareCaptionStyle{ruled}{labelfont=normalfont,labelsep=colon,strut=off} 
\usepackage{algorithm}
\usepackage[noend]{algorithmic}

\usepackage{newfloat}
\usepackage{listings}
\floatstyle{ruled}
\newfloat{listing}{tb}{lst}{}
\floatname{listing}{Listing}

\usepackage{graphicx}
\usepackage{amsmath}
\usepackage{amssymb}
\usepackage{amsthm}
\usepackage{syntax}
\usepackage{multirow}

\DeclareMathOperator*{\argmin}{arg\,min}
\title{Domain-Independent Dynamic Programming: \\
Generic State Space Search for Combinatorial Optimization}
\author{Ryo Kuroiwa, J. Christopher Beck}
\affiliations{
    Department of Mechanical and Industrial Engineering, University of Toronto, Toronto, Canada, ON M5S 3G8\\
    ryo.kuroiwa@mail.utoronto.ca, jcb@mie.utoronto.ca
}

\begin{document}

\maketitle

\begin{abstract}
    For combinatorial optimization problems, model-based approaches such as mixed-integer programming (MIP) and constraint programming (CP) aim to decouple modeling and solving a problem: the `holy grail' of declarative problem solving.
    We propose domain-independent dynamic programming (DIDP), a new model-based paradigm based on dynamic programming (DP).
    While DP is not new, it has typically been implemented as a problem-specific method.
    We propose Dynamic Programming Description Language (DyPDL), a formalism to define DP models, and develop Cost-Algebraic A* Solver for DyPDL (CAASDy), a generic solver for DyPDL using state space search.
    We formalize existing problem-specific DP and state space search methods for combinatorial optimization problems as DP models in DyPDL.
    Using CAASDy and commercial MIP and CP solvers, we experimentally compare the DP models with existing MIP and CP models, showing that, despite its nascent nature, CAASDy outperforms MIP and CP on a number of common problem classes.
\end{abstract}

\section{Introduction}
Combinatorial optimization is a central topic of artificial intelligence (AI) with many application fields including planning and scheduling.
In model-based approaches to such problems, users formulate problems as mathematical models and use generic solvers to solve the models.
While problem-specific algorithms can sometimes be more efficient, model-based approaches embody the quest for declarative general-purpose problem solving where it is sufficient to define the problem in order for it to be solved \cite{Freuder1997}.
While mixed-integer programming (MIP) and constraint programming (CP) are common in operations research (OR), domain-independent AI planning can also be considered a model-based approach.

Dynamic programming (DP) formulates a problem as a mathematical model using recursive equations with a state-based problem representation.
Although problem-specific DP achieves state-of-the-art performance on multiple problems, little work has considered DP as a generic technology.

In this paper, we propose \emph{domain-independent dynamic programming (DIDP)}, a new model-based paradigm for combinatorial optimization, based on DP.
Separating DP as a problem form from DP as an algorithm, we define Dynamic Programming Description Language (DyPDL), an algorithm-independent modeling formalism for DP.
We also develop Cost-Algebraic A* Solver for DyPDL (CAASDy), a DyPDL solver using a heuristic search algorithm.
While we adopt a relatively simple algorithm, the cost algebraic version of A* \cite{Edelkamp2005}, we demonstrate that it performs better than state-of-the-art MIP and CP approaches in four out of six problem classes.

\section{Dynamic Programming}
In DP, a problem is formulated as a \emph{state}, and the solution corresponds to a sequence of \emph{decisions}.
In a state $S$, one decision $t$ is made from a set of applicable decisions $\mathcal{T}(S)$, and $S$ is decomposed into a set of subproblems (states) $\mathcal{S}_t$.
The optimal cost of a problem is given by a value function $V$.
For a trivial subproblem $S$, $V(S)$ is defined by a constant.
Otherwise, $V(S)$ is defined by a function $F$ of a decision and costs of the subproblems in the following recursive equation:
\[
    V(S) = \min_{t \in \mathcal{T}(S)} F(t, \{ V(S') \mid S' \in \mathcal{S}_t \}).
\]
For maximization, $\min$ is replaced with $\max$.
The equation is usually solved by a problem-specific algorithm.

\section{DyPDL: A Modeling Formalism for DP}
Dynamic Programming Description Language (DyPDL) is a solver-independent formalism for a DP model.
DyPDL is inspired by AI planning formalisms such as STRIPS \cite{FikesN71}.
While domain-independent AI planning takes the `physics, not advice' approach, where a model contains only information necessary to define a problem, DyPDL allows a user to explicitly model implications of the definition.
Such is the standard convention in OR and commonly exploited in DP algorithms (e.g., \citeauthor{Dumas1995}~\shortcite{Dumas1995}).
DyPDL enables a user to express such information in a DP model.

\subsection{Example: TSPTW}
In a traveling salesperson problem with time windows (TSPTW), a set of customers $N = \{0,..,n\}$ is given.
A solution is a tour starting from the depot (index $0$), visiting each customer exactly once, and returning to the depot.
Visiting customer $j$ from $i$ incurs the travel time $c_{ij} \geq 0$.
In the beginning, $t=0$.
The visit to customer $i$ must be within a time window $[a_i, b_i]$.
Upon earlier arrival, waiting until $a_i$ is required.
The objective is to minimize the total travel time.

In the DP model proposed by \citeauthor{Dumas1995}~\shortcite{Dumas1995}, a state is a tuple of variables $\langle U, i, t \rangle$, which represents the set of unvisited customers, the current location, and the current time, respectively.
In this model, one customer is visited at each step.
The set of customers that can be visited next is $U' = \{ j \in U \mid t + c_{ij} \leq b_j\}$, and $t^j = \max\{ t + c_{ij}, a_j \}$ is the time when $j$ is visited from the current state.
Also, we use $c^*_{ij}$ as the shortest travel time from $i$ to $j$ ignoring time window constraints, which can be replaced with $c_{ij}$ when the triangle inequality holds.
    {\small
        \begin{align}
             & \text{compute } V(N \setminus \{ 0 \}, 0, 0) \label{eqn:tsptw:objective}                           \\
             & V(U, i, t) = \infty
            \quad \quad \quad \quad \quad \quad \quad \quad ~~
            \text{if } \exists j \in U, t + c^*_{ij} > b_j \label{eqn:tsptw:pruning}                              \\
             & V(U, i, t) = \begin{cases}
                                c_{i0}                                                  & \text{if } U = \emptyset    \\
                                \min_{j \in U'} c_{ij} + V(U \setminus \{ j \}, j, t^j) & \text{if } U \neq \emptyset
                            \end{cases} \label{eqn:tsptw:transitions} \\
             & V(U, i, t) \leq V(U, i, t')
            \quad \quad \quad \quad \quad \quad \quad \quad \quad \quad \quad ~~
            \text{if } t \leq t' \label{eqn:tsptw:dominance}                                                      \\
             & V(U, i, t) \geq 0. \label{eqn:tsptw:bound}
        \end{align}
    }%
Objective~\eqref{eqn:tsptw:objective} declares that the optimal cost is $V(N \setminus \{ 0 \},0,0)$, the cost to visit all customers starting from the depot with $t=0$.
In Equation~\eqref{eqn:tsptw:transitions}, the first line corresponds to returning to the depot from customer $i$, and the second line corresponds to visiting customer $j$ from $i$.
We assume that $V(U, i, t) = \infty$ if $U \neq \emptyset$ and $U' = \emptyset$.

Equation~\eqref{eqn:tsptw:pruning} and Inequalities~\eqref{eqn:tsptw:dominance} and \eqref{eqn:tsptw:bound} are not necessary to define a problem and are not present in the original DP model.
However, they were used algorithmically by \citeauthor{Dumas1995}~\shortcite{Dumas1995} as pruning rules;
when enumerating states, a state $\langle U, i, t \rangle$ is ignored if $\exists j \in U, t + c^*_{ij} > b_j$ since $j$ cannot be visited by its deadline.
A state $\langle U, i, t' \rangle$ is ignored if a state $\langle U, i, t \rangle$ with $t \leq t'$ is already considered because smaller $t$ leads to a better solution.
We formulate these pruning rules as Equation~\eqref{eqn:tsptw:pruning} and Inequality~\eqref{eqn:tsptw:dominance}.
In addition, we use the trivial lower bound of $0$ for $V$ in Inequality~\eqref{eqn:tsptw:bound}.

A state may satisfy multiple conditions in the model;
when $\exists j \in U, t + c^*_{ij} > b_j$ holds, $U \neq \emptyset$ also holds.
In such a case, we assume that the first condition defined is active.

\subsection{Formalism}
A DyPDL model is a tuple $\langle \mathcal{V}, S^0, \mathcal{K}, \mathcal{T}, \mathcal{B}, \mathcal{C}, h \rangle$, where $\mathcal{V}$ is a set of \emph{state variables}, $S^0$ is a \emph{target state}, $\mathcal{K}$ is a set of \emph{constants}, $\mathcal{T}$ is a set of \emph{transitions}, $\mathcal{B}$ is a set of \emph{base cases}, $\mathcal{C}$ is a set of \emph{state constraints}, and $h$ is a \emph{dual bound}.

\subsubsection{State Variables}
A state is defined by state variables, which can be \emph{element, set}, and \emph{numeric variables}.
An element or a set variable is associated with \emph{objects}, and the number of the objects is specified.
If there are $n$ objects, they are indexed from $0$ to $n-1$, and the values of element and set variables can be $i \in \{ 0,...,n-1\}$ and $M \subseteq \{ 0,...,n-1 \}$, respectively.
The value of a numeric variable is a real number.
In TSPTW, $U$ is a set variable, $i$ is an element variable, and $t$ is a numeric variable.
Objects $\{0, ..., n\}$ representing customers are associated with $U$ and $i$.

For an element or a numeric variable, a \emph{preference} for a greater or smaller value can be specified.
If the preference is specified for a variable, it is called a \emph{resource variable}.
When all other variables are the same in two states $S$ and $S'$, if each resource variable in $S$ is better than or equal to that of $S'$ according to its preference, then $V(S)$ is assumed to be better than $V(S')$.
In this case, we say that $S$ dominates $S'$, denoted by $S' \preceq S$.
In TSPTW, $\langle U, i, t' \rangle \preceq \langle U, i, t \rangle$ if $t \leq t'$ as defined in Inequality~\eqref{eqn:tsptw:dominance}.
This dominance is not necessary to define a model but can help solvers if present.

\subsubsection{Target State}
The target state $S^0$ must be specified as a full value assignment to state variables.
The objective of a DP model is to compute $V(S^0)$, the value of the target state.
In TSPTW, the target state is $\langle N \setminus \{ 0 \}, 0, 0 \rangle$ as in Objective~\eqref{eqn:tsptw:objective}.

\subsubsection{Constants}
A constant is a state-independent value.
Constants can be \emph{element, set, numeric}, and \emph{boolean constants}.
An element constant is a nonnegative integer representing the index of an object, and
a set constant is a set of nonnegative integers representing a set of the indices of objects.
A numeric constant is a real number and a boolean constant is a boolean value.
Multidimensional tables of constants can be defined and indexed by objects.
In TSPTW, $a_i$ and $b_i$ are constants in one-dimensional tables indexed by customer $i$, and $c_{ij}$ and $c^*_{ij}$ are constants in two-dimensional tables indexed by customers $i$ and $j$.

\subsubsection{Expressions}
Expressions are used in transitions, base cases, state constraints, and the dual bound to describe the computation of a value using the values of state variables and constants.
When an expression $e$ is evaluated given a state $S$, it returns a value $e(S)$.
Depending on the type of the returned value, we define \emph{element expressions}, \emph{set expressions}, \emph{numeric expressions}, and \emph{conditions}.

An element expression returns the index of an object.
It can refer to an element constant or variable and use arithmetic operations such as addition and subtraction on two element expressions.
In TSPTW, $j$ in Equation~\eqref{eqn:tsptw:transitions} is an element expression referring to the constant $j$.

A set expression returns a set of the indices of objects.
It can refer to a set constant or variable, add (remove) an element expression to (from) a set expression, and take the union/intersection/difference of two set expressions.
In TSPTW, $U \setminus \{ j \}$ in Equation~\eqref{eqn:tsptw:transitions} is a set expression removing $j$ from $U$.

A numeric expression returns a real number.
It can refer to a numeric constant or variable, use arithmetic operations, and take the cardinality of a set expression.
In TSPTW, $t^j = \max\{ t+c_{ij}, a_j \}$ in Equation~\eqref{eqn:tsptw:transitions} is a numeric expression.
It accesses a constant in a table, $c_{ij}$, using element expressions $i$ and $j$.
While $j$ refers to a constant, $i$ refers to a variable, so $c_{ij}$ depends on a state.
It is also possible to take the sum of numeric constants in a table using set expressions as shown below in the DP models for a simple assembly line balancing problem, bin packing, and graph-clear.

A condition returns a boolean value.
For a condition $c$ and a state $S$, we say $S \models c$ if $c(S) = \top$.
In addition to preconditions of transitions, base cases, and state constraints, a condition can be used to define an `if-then-else' expression, where one expression is returned if $S \models c$ and another is returned if not.
A condition can refer to a boolean constant, compare two element, numeric, or set expressions, and check whether an element is included in a set.
The conjunction and the disjunction of two conditions are also conditions.
In TSPTW, $\exists j \in U, t + c^*_{ij} > b_j$ in Equation~\eqref{eqn:tsptw:pruning} and $U = \emptyset$ in Equation~\eqref{eqn:tsptw:transitions} are conditions.

\subsubsection{Transitions}
A transition is a decision in DP and defines a recursive formula.
In DyPDL, we focus on problems where a state is transformed into another state by a decision, i.e., there is only one subproblem, so we call it a transition.
A transition $\tau$ is a 4-tuple $\langle \mathsf{eff}_\tau, \mathsf{cost}_\tau, \mathsf{pre}_\tau, \mathsf{forced}_\tau \rangle$ where the set of \emph{effects} $\mathsf{eff}_\tau$ describes how to transform a state to another state, the \emph{cost expression} $\mathsf{cost}_\tau$ describes how to compute $V(S)$, the set of \emph{preconditions} $\mathsf{pre}_\tau$ describes when the transition is applicable, and $\mathsf{forced}_\tau$ is a boolean indicating whether it is a \emph{forced transition}.
In TSPTW, transitions are defined in Equation~\eqref{eqn:tsptw:transitions}.

For each state variable, $\mathsf{eff}_\tau$ defines an expression to update it.
By $S[\![\tau]\!]$, we denote the updated state by transition $\tau$ from state $S$.
In TSPTW, in Equation~\eqref{eqn:tsptw:transitions}, the set expression $U \setminus \{ j \}$ updates the state variable $U$, the element expression $j$ updates the state variable $i$, and the numeric expression $\max\{t+c_{ij}, a_j\}$ updates the numeric variable $t$.

The cost expression $\mathsf{cost}_\tau$ is a numeric expression describing the computation of $V(S)$.
In addition to variables and constants, it can use $V(S[\![\tau]\!])$, the value of the transformed state.
By $\mathsf{cost}_\tau(V(S[\![\tau]\!]), S)$, we denote the value of $\mathsf{cost}_\tau$ given $V(S[\![\tau]\!])$ and $S$.
In TSPTW, the cost expression of a transition for $j \in U$ in Equation~\eqref{eqn:tsptw:transitions} is $c_{ij} + V(S[\![\tau]\!])$.

A precondition in $\mathsf{pre}_\tau$ is a condition, and the transition $\tau$ is applicable in a state $S$ only if $S \models p$ for each $p \in \mathsf{pre}_\tau$, denoted by $S \models \mathsf{pre}_\tau$.
In TSPTW, $t + c_{ij} \leq b_j$ and $j \in U$ are preconditions of a transition defined in Equation~\eqref{eqn:tsptw:transitions}.
If $\mathsf{forced}_\tau = \top$, $\tau$ is a forced transition.
When the preconditions of a forced transition are satisfied, all other transitions are ignored.
Let $\mathcal{T}_f = \{ \tau \in \mathcal{T} \mid \mathsf{forced}_\tau \}$ be the set of forced transitions.
The set of applicable transitions in $S$ is
    {\small
        \[
            \mathcal{T}(S) = \begin{cases}
                \{ \tau \}                                                  & \text{if } \exists \tau \in \mathcal{T}_f, S \models \mathsf{pre}_\tau \\
                \{ \tau \in \mathcal{T} \mid S \models \mathsf{pre}_\tau \} & \text{otherwise.}
            \end{cases}
        \]
    }%
If multiple forced transitions are applicable, the first one defined is used.
A forced transition can be used to break symmetry as shown below in the DP model for bin packing.

In minimization, $V(S)$ is computed as the minimum over all applicable transitions, i.e., $V(S) = \min_{\tau \in \mathcal{T}(S)} \mathsf{cost}_\tau(V(S[\![\tau]\!]), S)$.
When $\mathcal{T}(S) = \emptyset$, then $V(S) = \infty$ is assumed.
In maximization, $V(S) = \max_{\tau \in \mathcal{T}(S)} \mathsf{cost}_\tau(V(S[\![\tau]\!]), S)$, and $V(S) = - \infty$ is assumed if $\mathcal{T}(S) = \emptyset$.
In TSPTW, $V(S) = \min_{j \in U'} c_{ij} + V(U \setminus \{ j \},j,t^j)$ as in the second line of Equation~\eqref{eqn:tsptw:transitions}.

\subsubsection{Base Cases}
A base case is defined as a set of conditions to terminate the recursion.
If $S \models \beta$ for every $\beta$ in a base case $B$, denoted by $S \models B$, then $V(S) = 0$.
We call such a state a \emph{base state}.
In TSPTW, since $V(U, i, t) = c_{i0}$ if $U = \emptyset$ in the first line of Equation~\eqref{eqn:tsptw:transitions}, we use a base case $\{ U = \emptyset, i = 0 \}$ and introduce a transition with cost expression $c_{i0} + V(U, 0, t+c_{i0})$ and preconditions $U = \emptyset$ and $i \neq 0$.

\subsubsection{State Constraints}
A state constraint is a condition that must be satisfied by all states.
If a state does not satisfy a state constraint, it can be immediately discarded.
In TSPTW, $V(U, i, t) = \infty$ if $\exists j \in U, t + c^*_{ij} > b_j$ (Equation~\eqref{eqn:tsptw:pruning}) is implemented as a state constraint $\forall j \in U, t + c^*_{ij} \leq b_j$.

\subsubsection{Dual Bound}
The dual bound $h$ is defined as a numeric expression, and $h(S)$ must be the lower (upper) bound on $V(S)$ for minimization (maximization).
The dual bound is not required but can be exploited by a solver.
In TSPTW, Inequality~\eqref{eqn:tsptw:bound} defines a dual bound, $h(S) = 0$ for all $S$.

\subsection{DyPDL Example for TSPTW}
Table~\ref{tab:dypdl} presents the DyPDL representation of the DP model for TSPTW in \eqref{eqn:tsptw:objective}--\eqref{eqn:tsptw:bound}.
The transition to visit customer $j$ is defined for all $j \in N$.
There are no forced transitions, i.e., $\mathsf{forced}_\tau = \bot$ for all $\tau \in \mathcal{T}$.

\begin{table}
    \small
    \tabcolsep=0.48em
    \begin{tabular}{l|lll}
        $\mathcal{V}$              & Type                                                                        & Objects                     & Preference            \\
        \hline
        $U$                        & set                                                                         & customers $N$               &                       \\
        $i$                        & element                                                                     & customers $N$               &                       \\
        $t$                        & numeric                                                                     &                             & less                  \\
        \hline
        \hline
        $\mathcal{K}$              & Type                                                                        & Indices                                             \\
        \hline
        $a_j$                      & numeric                                                                     & $j \in N$                                           \\
        $b_j$                      & numeric                                                                     & $j \in N$                                           \\
        $c_{jk}$                   & numeric                                                                     & $j,k \in N$                                         \\
        $c^*_{jk}$                 & numeric                                                                     & $j,k \in N$                                         \\
        \hline
        \hline
        $S^0$                      & \multicolumn{3}{l}{$\langle U = N \setminus \{ 0 \}, i = 0, t = 0 \rangle$}                                                       \\
        $\mathcal{B}$              & \multicolumn{3}{l}{$\{ \{ U = \emptyset, i = 0 \} \}$}                                                                            \\
        $\mathcal{C}$              & \multicolumn{3}{l}{$\{ \forall j \in U, t + c^*_{ij} \leq b_j \}$}                                                                \\
        $h$                        & \multicolumn{3}{l}{$0$}                                                                                                           \\
        \hline
        \hline
        $\mathcal{T}$              & $\mathsf{eff}$                                                              & $\mathsf{cost}$             & $\mathsf{pre}$        \\
        \hline
        \multirow{3}{*}{visit $j$} & $U \leftarrow U \setminus \{ j \}$                                          & $c_{ij} + V(S[\![\tau]\!])$ & $j \in U$             \\
                                   & $i \leftarrow j$                                                            &                             & $t + c_{ij} \leq b_j$ \\ & $t \leftarrow \max\{ t + c_{ij}, a_j \}$ &                  & \\
        \hline
        \multirow{2}{*}{return}    & $i \leftarrow 0$                                                            & $c_{i0} + V(S[\![\tau]\!])$ & $U = \emptyset$       \\
                                   & $t \leftarrow t + c_{i0}$                                                   &                             & $i \neq 0$
    \end{tabular}
    \caption{DyPDL representation of the DP model for TSPTW. No forced transition exists in this model.}
    \label{tab:dypdl}
\end{table}

\subsection{YAML-DyPDL: An Implementation of DyPDL}
We propose YAML-DyPDL, an implementation of DyPDL based on the YAML data format,\footnote{https://yaml.org/} inspired by PDDL \cite{Ghallab1998}.
A problem instance is represented by \emph{domain} and \emph{problem files}.
While a domain file can be shared by multiple instances of the same problem, a problem file is specific to one problem instance.
A domain file defines objects, state variables, state constraints, base cases, transitions, and the dual bound and declares tables of constants.
A problem file defines the number of objects, the target state, and the values of the constants in the tables.
In addition, state constraints, base cases, transitions, and dual bounds can be also defined in a problem file.
We show an example of a domain file in Listing~\ref{lst:domain} and a problem file in Listing~\ref{lst:problem}, which correspond to the DP model for TSPTW in \eqref{eqn:tsptw:objective}--\eqref{eqn:tsptw:bound}.
In this example, assuming that the triangle inequality holds, $c^*_{ij}$ is replaced with $c_{ij}$, and $t + c_{ij} \leq b_j$ is removed from preconditions of the transition to visit $j$ because it is ensured by the state constraint $\forall j \in U, t + c_{ij} \leq b_j$.

In YAML, key-value pairs are defined, where values can be numeric values, strings, lists of values, and key-value pairs.
Objects, state variables, a target state, and tables of constants are defined directly using key-value pairs and lists.
In addition, \texttt{reduce: min} in line~35 of the domain file specifies to minimize the objective.
In lines~3--12, state variables with names \texttt{U}, \texttt{i}, and \texttt{t} are defined, corresponding to $U$, $i$, and $t$, respectively.
Tables \texttt{a}, \texttt{b}, and \texttt{c} in lines~13--26 correspond to $a$, $b$, and $c$, respectively.
Numeric variables, constants, and the domain of the value function are either of \texttt{integer} or \texttt{continuous}, corresponding to integer and continuous values.
In the definitions of transitions, base cases, state constraints, and the dual bound, an expression is written as a string following a LISP-like syntax.
In an expression, a constant in a table is accessed by its name and indices, e.g., \texttt{(c i j)} in line~28 corresponds to a constant $c_{ij}$, where \texttt{i} is a state variable and \texttt{j} is a constant.

A quantifier \texttt{forall} is used to define the conjunction of conditions that are only different in element constants, which have the same object type, or are included in the same set variable.
In lines~28--31, \texttt{forall} is used in the state constraint with the set variable \texttt{U}, and the element constant \texttt{j} is used in the expression, corresponding to $\forall j \in U$.
Similarly, multiple transitions can be defined with \texttt{parameters}.
In the example, to define transitions in the second line of Equation~\eqref{eqn:tsptw:transitions}, only one definition parameterized by \texttt{j} in \texttt{U} is used in lines~38--46.
Since \texttt{U} is a state variable, a precondition \texttt{(is_in j U)}, corresponding to $j \in U$, is assumed.

Note that implementations of DyPDL are not necessarily restricted to YAML-DyPDL, which adopts a style common in AI planning.
Developing and improving interfaces for DyPDL is part of our future work.
For example, a Python library will be useful for OR researchers and practitioners.

\begin{listing}[!tb]
\begin{lstlisting}
objects:
  - customer
state_variables:
  - name: U
    type: set
    object: customer
  - name: i
    type: element
    object: customer
  - name: t
    type: integer
    preference: less
tables:
  - name: a
    type: integer
    args:
      - customer
  - name: b
    type: integer
    args:
      - customer
  - name: c
    type: integer
    args:
      - customer
      - customer
constraints:
  - condition: (<= (+ t (c i j)) (b j))
    forall:
      - name: j
        object: U
base_cases:
  - - (is_empty U)
    - (= i 0)
reduce: min
cost_type: integer
transitions:
  - name: visit
    parameters:
      - name: j
        object: U
    effect:
      U: (remove j U)
      i: j
      t: (max (+ t (c i j)) (a j))
    cost: (+ cost (c i j))
  - name: return
    preconditions:
      - (is_empty U)
      - (!= i 0)
    effect:
      i: 0
      t: (+ t (c i 0))
    cost: (+ cost (c i 0))
dual_bounds:
  - 0
\end{lstlisting}
    \caption{YAML-DyPDL domain file for TSPTW.}
    \label{lst:domain}
\end{listing}

\begin{listing}[!tb]
\begin{lstlisting}
object_numbers:
  customer: 4
target:
  U: [1, 2, 3]
  i: 0
  t: 0
table_values:
  a: { 1: 5, 2: 0, 3: 8 }
  b: { 1: 16, 2: 10, 3: 14 }
  c:
    {
      [0, 1]: 3, [0, 2]: 4, [0, 3]: 5,
      [1, 0]: 3, [1, 2]: 5, [1, 3]: 4,
      [2, 0]: 4, [2, 1]: 5, [2, 3]: 3,
      [3, 0]: 5, [3, 1]: 4, [3, 2]: 3,
    }
\end{lstlisting}
    \caption{YAML-DyPDL problem file for TSPTW.}
    \label{lst:problem}
\end{listing}

\section{CAASDy: A State Space Search Solver for DP}
While various approaches can be applied to solve DyPDL, for our prototype solver, we adopt cost-algebraic heuristic search \cite{Edelkamp2005}.
A DyPDL problem can be considered a graph search problem, where nodes correspond to states, edges correspond to transitions, and a solution corresponds to a path from $S^0$ to a base state.
To compute the cost of a solution, we need to evaluate the cost expressions of the transitions backward from a base state to the target state.
A naive approach is to perform recursion according to the recursive equations while memoizing all encountered states.
However, if the cost expressions of transitions $\tau \in \mathcal{T}$ are in the form of $e_\tau(S) \times V(S[\![\tau]\!])$, where $e_\tau$ is a numeric expression and $\times$ is a binary operator, and the cost-algebra conditions are satisfied, the optimal solution can be computed by cost-algebraic search algorithms, generalized versions of shortest path algorithms such as A* \cite{Hart1968}.
For example, if the cost expression is in the form of $e_\tau(S) + V(S[\![\tau]\!])$ with $e_\tau(S) \geq 0$ for all $S$, and the objective is minimization, the optimal solution corresponds to the shortest path in a graph where the weight of edge $(S, S[\![\tau]\!])$ is $e_\tau(S)$.
TSPTW is such an example since the cost expression of each transition is $c_{ij} + V(S[\![\tau]\!])$.
In addition, a minimization problem with cost expressions in the form of $\max\{ e_\tau(S), V(S[\![\tau]\!]) \}$ with $e_\tau(S) \geq 0$ also satisfies the property of cost-algebra, as proved in the appendix.

\begin{algorithm}[!tb]
    \begin{algorithmic}[1]
        \STATE $g(S^0) \leftarrow 0, f(S^0) \leftarrow h(S^0), O, G \leftarrow \{ S^0 \}$.
        \WHILE{$O \neq \emptyset$}
        \STATE Let $S \in \argmin_{S \in O} f(S)$.
        \STATE $O \leftarrow O \setminus \{ S \}$. \label{alg:astar:pop}
        \IF{$\exists B \in \mathcal{B}, S \models B$} \RETURN $g(S)$. \ENDIF
        \FORALL{$\tau \in \mathcal{T}(S)$}
        \IF{$\forall c \in \mathcal{C}, S[\![\tau]\!] \models c$} \label{alg:astar:constraints}
        \STATE $g^\tau \leftarrow g(S) \times e_\tau(S)$. \label{alg:astar:g}
        \IF{$\nexists S' \in G, S[\![\tau]\!] \preceq S' \land g^\tau \geq g(S') $} \label{alg:astar:dominance}
        \STATE $g(S[\![\tau]\!]) \leftarrow g^\tau, f(S[\![\tau]\!]) \leftarrow g^\tau \times h(S[\![\tau]\!])$. \label{alg:astar:f}
        \STATE $G \leftarrow G \cup \{ S[\![\tau]\!] \}, O \leftarrow O \cup \{ S[\![\tau]\!] \}$.
        \ENDIF \ENDIF
        \ENDFOR
        \ENDWHILE
        \RETURN $\infty$.
    \end{algorithmic}
    \caption{Cost-Algebraic A* for DyPDL}
    \label{alg:astar}
\end{algorithm}

We adopt the cost-algebraic version of A* \cite{Edelkamp2005} and name our solver Cost-Algebraic A* Solver for DyPDL (CAASDy).
To solve a problem optimally, A* uses an admissible heuristic function, which computes a lower bound of the shortest path cost from a node.
Cost-algebraic A* also uses a heuristic function, which computes a lower (upper) bound for minimization (maximization).
In CAASDy, we do not use any hand-coded heuristic function.
Instead, CAASDy just uses the dual bound $h$ defined by a user in a DyPDL model as a heuristic function.
In TSPTW, the trivial lower bound $V(S) \geq 0$ is defined in Inequality~\eqref{eqn:tsptw:bound}, so $h(S) = 0$ for all $S$ is used as a heuristic function.

We show the pseudo-code of CAASDy in Algorithm~\ref{alg:astar}.
While we focus on minimization in the pseudo-code, it can be easily adapted to maximization as long as the cost-algebra conditions are satisfied.
For each state $S$, the path cost from the target state $g(S)$ (the $g$-value), the heuristic value $h(S)$ (the $h$-value), and the priority $f(S) = g(S) \times h(S)$ (the $f$-value) are maintained.
Recall that $\times$ is the binary operator used in the cost expressions satisfying the cost-algebra conditions, e.g., $+$ and $\max$.
The open list $O$ is the set of candidate states to search, and $G$ stores all generated states.
In line~\ref{alg:astar:pop}, a state $S$ minimizing $f(S)$ is removed from $O$.
We select the state minimizing $h(S)$ if there are multiple candidates.
As discussed above, $h$ is defined as the dual bound in a DyPDL model.
If there are multiple states with the same $f(S)$ and $h(S)$, the tie-breaking depends on the binary heap implementation of $O$.
In line~\ref{alg:astar:constraints}, a state is pruned if a state constraint is not satisfied.
As in line~\ref{alg:astar:dominance}, a state is inserted into $O$ only if there is no dominating state having an equal or smaller $g$-value in $G$.
In the implementation, we maintain a hash table, where a key is the state variable values excluding resource variables, and a value is a list of states and their $g$-values.
For each generated state, we check if its key exists in the hash table.
If it exists, we compare the $g$-values and the resource variables of the state with each state in the list.

\section{DP Models for Combinatorial Optimization}
We present DP models for combinatorial optimization problems that can be represented in DyPDL.
While we show recursive equations because they are succinct and easy to understand, the DyPDL representations and YAML-DyPDL files are provided in the appendix.

\subsubsection{CVRP}
In a capacitated vehicle routing problem (CVRP), customers $N = \{ 0,...,n \}$, where $0$ is the depot, are given, and each customer $i \in N \setminus \{ 0 \}$ has the demand $d_i$.
A solution is to visit each customer in $N \setminus \{ 0 \}$ exactly once using $m$ vehicles, which start from and return to the depot.
The sum of demands of customers visited by a single vehicle must be less than or equal to the capacity $q$.
Visiting customer $j$ from $i$ incurs the travel time $c_{ij} \geq 0$, and the objective is to minimize the total travel time.

We formulate the DP model based on the giant-tour representation \cite{Gromicho2012}.
We sequentially construct tours for the $m$ vehicles.
Let $U$ be a set variable representing unvisited customers, $i$ be an element variable representing the current location, $l$ be a numeric variable representing the current load, and $k$ be a numeric variable representing the number of used vehicles.
Both $l$ and $k$ are resource variables where less is preferred.
At each step, one customer is visited by the current vehicle or a new vehicle.
When a new vehicle is used, the customer is visited via the depot, $l$ is reset, and $k$ is increased.
Let $U' = \{ j \in U \mid l + d_j \leq q \}$ be the set of customers that can be visited next by the current vehicle, and $c'_{ij} = c_{i0} + c_{0j} $ be a numeric constant representing the travel time from $i$ to $j$ via the depot.
    { \small
        \begin{align*}
             & \text{compute } V(N \setminus \{ 0 \}, 0, 0, 1)                                                                                        \\
             & V(\emptyset, i, l, k) =                c_{i0}                                                                                          \\
             & V(U, i, l, k) =                   \min \left\{ \begin{array}{l}
                                                                  \min_{j \in U'} c_{ij} + V(U \setminus \{ j \}, j, l + d_j, k) \\
                                                                  \min_{j \in U} c'_{ij} + V(U \setminus \{ j \}, j, d_j, k + 1) \\
                                                              \end{array}\right.                          \\
             & \quad \quad \quad \quad \quad \quad \quad \quad \quad \quad \quad \quad \quad \quad \quad \quad \quad \quad \quad \quad \quad \quad ~~
            \text{ if } k < m                                                                                                                         \\
             & V(U, i, l, k) = \min_{j \in U'} c_{ij} + V(U \setminus \{ j \}, j, l + d_j, k) \quad \quad \text{ if } k = m                           \\
             & V(U, i, l, k) \leq V(U, i, l', k')
            \quad \quad \quad \quad \quad \quad \quad ~~
            \text{if } l \leq l' \land k \leq k'                                                                                                      \\
             & V(U, i, l, k) \geq 0.
        \end{align*}
    }

\subsubsection{SALBP-1}
In a simple assembly line balancing problem (SALBP), tasks $N = \{0,...,n-1\}$ are given, and each task $i$ has processing time $t_i$ and predecessors $P_i \subset N$.
A solution assigns tasks to a totally ordered set of stations so that the sum of processing times at each station does not exceed the cycle time $c$, and all tasks in $P_i$ are scheduled in the same station as $i$ or an earlier station.
We focus on minimizing the number of stations, which is called SALBP-1, for which branch-bound-and-remember (BB\&R), a state space search algorithm, is a state-of-the-art exact method \cite{Morrison2014}.
We formulate a DP model inspired by BB\&R.
Let $U$ be a set variable representing unscheduled tasks and $r$ be a numeric variable representing the remaining time in the current station.
Since having more remaining time leads to a better solution if the sets of unscheduled tasks are the same, $r$ is a resource variable where more is preferred.
Let $U' = \{ i \in U \mid P_i \cap U = \emptyset \land r \geq t_i \}$ be tasks that can be assigned to the current station.
At each step, one task is assigned to the current station from $U'$, or a new station is opened when $U' = \emptyset$, which is called the maximal load pruning rule in the literature.

The model has a dual bound based on lower bounds of the bin packing problem obtained by ignoring predecessors (see the appendix).
We define tables of numeric constants $w^2$, $w'^2$, and $w^3$ indexed by a task $i$, whose values depend on $t_i$.
\begin{center}
    \small
    \begin{tabular}{c|ccc}
        $t_i$    & $(0, c/2)$ & $c/2$ & $(c/2, c]$ \\
        \hline
        $w^2_i$  & $0$        & $0$   & $1$        \\
        $w'^2_i$ & $0$        & $1/2$ & $0$        \\
    \end{tabular}
    \begin{tabular}{c|ccccc}
        $t_i$   & $(0, c/3)$ & $c/3$ & $(c/3, c/2)$ & $2c/3$ & $(2c/3, c]$ \\
        \hline
        $w^3_i$ & $0$        & $1/3$ & $1/2$        & $2/3$  & $1$
    \end{tabular}
\end{center}
In addition, we use an `if-then-else' expression $l^2$, which returns $1$ if $r \geq c/2$ and $0$ otherwise.
Similarly, an expression $l^3$ returns $1$ if $r \geq c/3$ and $0$ otherwise.
In YAML-DyPDL, $l^2$ is written as \texttt{(if (>= r (/ c 2.0)) 1 0)}.
{\small
\begin{align*}
     & \text{compute } V(N, 0)                                                                    \\
     & V(U, r) = \begin{cases}
                     0                                             & \text{if } U = \emptyset     \\
                     \min_{i \in U'} V(U \setminus \{ i \}, r-t_i) & \text{if } U' \neq \emptyset \\
                     1 + V(U, c)                                   & \text{if } U' = \emptyset
                 \end{cases}     \\
     & V(U, r) \leq V(U, r')
    \quad \quad \quad \quad \quad \quad \quad \quad \quad ~~~
    \text{if } r \geq r'                                                                          \\
     & V(U, r) \geq \max \left\{ \begin{array}{l}
                                     \lceil (\sum_{i \in U} t_i - r) / c \rceil                       \\
                                     \sum_{i \in U} w^2_i + \lceil \sum_{i \in U} w'^2_i \rceil - l^2 \\
                                     \lceil \sum_{i \in U} w^3_i \rceil - l^3                         \\
                                 \end{array} \right.
\end{align*}
}%
In this model the sum of constants in a table, e.g., $\sum_{i \in U}t_i$, is used.
In YAML-DyPDL, it is expressed as \texttt{(sum t U)}.

\subsubsection{Bin Packing}
A bin packing problem is the same as SALBP-1 except that a task has no predecessors.
We call a task an item and a station a bin and pack an item in a bin instead of assigning a task to a station.
We adapt the DP model for SALBP-1 to bin packing.
In addition to $U$ and $r$, the model has an element resource variable $k$ representing the number of used bins, where less is preferred.
The model breaks symmetry by packing item $i$ in the $i$-th or an earlier bin.
Thus, $U^1 = \{ i \in U \mid r \geq t_i \land i + 1 \geq k \}$ represents items that can be packed in the current bin.
When $U^1 = \emptyset$, then a new bin is opened, and any item in $U^2 = \{ i \in U \mid i \geq k \}$ can be packed.
The model also breaks symmetry here by selecting an arbitrary item in $U^2$, implemented as a forced transition.
    {\small
        \begin{align*}
             & \text{compute } V(N, 0, 0)                                                                     \\
             & V(U, r, k) = \begin{cases}
                                0                                                 & \text{if } U = \emptyset      \\
                                \min_{i \in U^1} V(U \setminus \{ i \}, r-t_i, k) & \text{if } U^1 \neq \emptyset \\
                                1 + V(U \setminus \{ i \}, c-t_i, k + 1)          & \text{if } \exists i \in U^2  \\
                                \infty                                            & \text{otherwise}
                            \end{cases} \\
             & V(U, r, k) \leq V(U, r', k')
            \quad \quad \quad \quad \quad \quad ~
            \text{if } r \geq r' \land k \leq k'                                                              \\
             & V(U, r, k) \geq  \max \left\{ \begin{array}{l}
                                                 \lceil (\sum_{i \in U} t_i - r) / c \rceil                       \\
                                                 \sum_{i \in U} w^2_i + \lceil \sum_{i \in U} w'^2_i \rceil - l^2 \\
                                                 \lceil \sum_{i \in U} w^3_i \rceil - l^3                         \\
                                             \end{array} \right.
        \end{align*}
    }

\subsubsection{MOSP}
In the minimization of open stacks problem (MOSP) \cite{Yuen1995}, customers $C = \{0,...,n-1\}$ and products $P = \{0,...,m-1\}$ are given, and each customer $c$ orders products $P_c \subseteq P$.
A solution is a sequence in which products are produced.
When producing product $i$, a stack for customer $c$ with $i \in P_c$ is opened, and it is closed when all of $P_c$ are produced.
The objective is to minimize the maximum number of open stacks at a time.

For MOSP, customer search is a state-of-the-art exact method \cite{Chu2009}.
It searches for an order of customers to close stacks, from which the order of products is determined;
for each customer $c$, all products ordered by $c$ and not yet produced are consecutively produced in an arbitrary order.
We formulate customer search as a DP model.
A set variable $R$ represents customers whose stacks are not closed, and $O$ represents customers whose stacks have been opened.
Let $N_c = \{ c' \in C \mid P_c \cap P_{c'} \neq \emptyset \}$ be a set constant representing customers that order the same product as $c$.
    {\small
        \begin{align*}
             & \text{compute } V(C, \emptyset)                                               \\
             & V(R, O) = \begin{cases}
                             0 & \text{if } R = \emptyset \\
                             \min\limits_{c \in R} \max\left\{ \begin{array}{l}
                                                      V(R \setminus \{ c \}, O \cup N_c) \\
                                                      |(O \cap R) \cup (N_c \setminus O)|
                                                  \end{array} \right.
                         \end{cases} \\
             & V(R, O) \geq 0
        \end{align*}
    }

\subsubsection{Graph-Clear}
In a graph-clear problem \cite{Kolling2007}, an undirected graph $(N, E)$ with the node weight $a_i$ for $i \in N$ and the edge weight $b_{ij}$ for $\{ i, j \} \in E$ is given.
In the beginning, all nodes are contaminated.
In each step, one node can be made clean by sweeping it using $a_i$ robots and blocking each edge $\{ i, j \}$ using $b_{ij}$ robots.
However, while sweeping a node, an already swept node becomes contaminated if it is connected by a path of unblocked edges to a contaminated node.
The optimal solution minimizes the maximum number of robots per step to make all nodes clean.

Previous work \cite{Morin2018} developed a state-based formula as the basis for MIP and CP models, but no DP model was defined.
Here, we propose such a model.
A set variable $C$ represents swept nodes, and one node in $\overline{C} = N \setminus C$ is swept at each step.
Weights $a_i$ and $b_{ij}$ are defined as numeric constants, assuming that $b_{ij} = 0$ if $\{ i, j \} \notin E$, and $N$ is defined as a set constant.
    {\small \begin{align*}
             & \text{compute } V(\emptyset)                                                                          \\
             & V(C) = \begin{cases}
                          0
                          \quad \quad \quad \quad \quad \quad \quad \quad \quad \quad \quad \quad \quad \quad \quad \quad
                          \text{if } C = N \\
                          \min\limits_{c \in \overline{C}} \max\left\{ \begin{array}{l}
                                                                 V(C \cup \{ c \}) \\
                                                                 a_c + \sum\limits_{i \in N} b_{ci} + \sum\limits_{i \in C}\sum\limits_{j \in \overline{C} \setminus \{ c \}} b_{ij}
                                                             \end{array} \right.
                      \end{cases} \\
             & V(C) \geq 0.
        \end{align*}}%
The model takes the sum of $b_{ij}$ over all combinations of $i \in C$ and $j \in \overline{C} \setminus \{ c \}$.
In YAML-DyPDL, it is described as \texttt{(sum b C (remove c \textasciitilde C))}.

\section{Experimental Evaluation}
We experimentally compare DP, MIP, and CP models for the problems presented above.
Most MIP and CP models are from the literature though we present new models in the appendix when they achieve superior performance to the literature.
We use CAASDy as a solver for all the DP models.
We use Gurobi Optimizer 9.5.1 for MIP and CP Optimizer from CPLEX Optimization Studio 22.1.0 for CP.
As CAASDy is not an anytime solver, i.e., the first found solution is the optimal solution, we evaluate the number of instances solved to optimality within time and memory limits.

We implement the YAML-DyPDL parser and CAASDy in Rust 1.62.1.\footnote{\url{https://github.com/domain-independent-dp/didp-rs}}
Problem instances are transformed from their benchmark format into YAML-DyPDL by a Python 3.10.4 script and passed to CAASDy.
We also use Python 3.10.4 to implement MIP and CP models.
We run all experiments on a machine running Ubuntu 22.04 with an Intel Core i7 11700 processor using GNU Parallel \cite{GNUParallel}.
For each instance, we use a single thread with a 30-minute time and 8 GB memory limit.
We show the results in Table~\ref{tab:result}.
In the last row, we present the ratio of optimally solved instances in each problem class averaged over all problem classes.
DP solves the largest ratio of instances on average.
Overall problems, we observe that if DP fails to solve an instance, it is due to the memory limit.

\begin{table}[t!]
    \small
    \centering
    \begin{tabular}{l|rrr}
    TSPTW                               & MIP          & CP            & DP            \\
    \hline
    Dumas (135)                         & 121          & 36            & \textbf{135}  \\
    GDE (130)                           & 71           & 4             & \textbf{77}   \\
    OT (25)                             & 0            & 0             & 0             \\
    AFG (50)                            & 33           & 7             & \textbf{45}   \\
    \hline
    Total (340)                         & 225          & 47            & \textbf{257}  \\
    \hline
    \hline
    CVRP                                & MIP          & CP            & DP            \\
    \hline
    A, B, E, F, P (90)                  & \textbf{26}  & 0             & 4             \\
    \hline
    \hline
    SALBP-1                             & MIP          & CP            & DP            \\
    \hline
    Small (525)                         & \textbf{525} & \textbf{525}  & \textbf{525}  \\
    Medium (525)                        & \textbf{518} & 501           & 509           \\
    Large (525)                         & 318          & 404           & \textbf{414}  \\
    Very large (525)                    & 0            & 155           & \textbf{204}  \\
    \hline
    Total (2100)                        & 1360         & 1585          & \textbf{1652} \\
    \hline
    \hline
    Bin Packing                         & MIP          & CP            & DP            \\
    \hline
    Falkenauer U (80)                   & 25           & \textbf{36}   & 33            \\
    Falkenauer T (80)                   & 37           & \textbf{56}   & 27            \\
    Scholl 1 (720)                      & \textbf{605} & 533           & 517           \\
    Scholl 2 (480)                      & 354          & \textbf{445}  & 335           \\
    Scholl 3 (10)                       & 0            & \textbf{1}    & 0             \\
    W\"{a}scher (17)                    & 2            & \textbf{10}   & \textbf{10}   \\
    Schwerin 1 (100)                    & 80           & \textbf{96}   & 0             \\
    Schwerin 2 (100)                    & 54           & \textbf{61}   & 0             \\
    Hard 28 (28)                        & \textbf{0}   & \textbf{0}    & \textbf{0}    \\
    \hline
    Total (1615)                        & 1157         & \textbf{1238} & 922           \\
    \hline
    \hline
    MOSP                                & MIP          & CP            & DP            \\
    \hline
    Constraint Modelling Challenge (46) & 41           & \textbf{44}   & \textbf{44}   \\
    SCOOP Project (24)                  & 8            & \textbf{23}   & 16            \\
    Faggioli and Bentivoglio (300)      & 130          & \textbf{300}  & 298           \\
    Chu and Stuckey (200)               & 44           & 70            & \textbf{125}  \\
    \hline
    Total (570)                         & 223          & 437           & \textbf{483}  \\
    \hline\hline
    Graph-Clear                         & MIP          & CP            & DP            \\
    \hline
    Planar (60)                         & 16           & 1             & \textbf{45}   \\
    Random (75)                         & 8            & 3             & \textbf{31}   \\
    \hline
    Total (135)                         & 24           & 4             & \textbf{76}   \\
    \hline
    \hline
    Average ratio                       & 0.48         & 0.41          & \textbf{0.59}
\end{tabular}
    \caption{
        Number of instances solved to optimality.
        `Average ratio' is the ratio of optimally solved instances in each problem class averaged over all problem classes.
    }
    \label{tab:result}
\end{table}

\subsubsection{TSPTW}
We use four benchmark sets, Dumas \cite{Dumas1995}, GDE \cite{Gendreau1998}, OT \cite{Ohlmann2007}, and AFG \cite{Ascheuer1995}.
In the DP model, as the travel time satisfies the triangle inequality, we replace $c^*_{ij}$ with $c_{ij}$.
For MIP, we use Formulation (1) proposed by \citeauthor{Hungerlander2018}~\shortcite{Hungerlander2018}.
When there are zero-cost edges, flow-based subtour elimination constraints \cite{Gavish1978} are added.
We adapt a CP model for a single machine scheduling problem with time windows \cite{Booth2016} to TSPTW, where an interval variable represents the time to visit a customer.
We change the objective to the sum of travel costs and add a $\mathsf{First}$ constraint ensuring that the depot is visited first.
DP solves more instances than MIP and CP benefiting from pruning based on time windows.

\subsubsection{CVRP}
We use A, B, E, F, and P instances from CVRPLIB \cite{Uchoa2017} because they have at most 100 customers.
The travel time is symmetric in these instances.
We use a MIP model proposed by \citeauthor{Gadegaard2021}~\shortcite{Gadegaard2021} and a CP model proposed by \citeauthor{Saadaoui2019}~\shortcite{Saadaoui2019}.
MIP solves more instances than DP.
Since there is no efficient pruning method unlike TSPTW, CAASDy suffers from an increasing branching factor with the number of customers and quickly runs out of memory.
CP does not solve any instances to optimality.

\subsubsection{SALBP-1}
We use the benchmark set proposed by \citeauthor{Morrison2014}~\shortcite{Morrison2014}.
For MIP, we use the NF4 formulation \cite{Ritt2018}.
We use a CP model proposed by \citeauthor{Bukchin2018}~\shortcite{Bukchin2018} but implement it using the global constraint $\mathsf{Pack}$ in CP Optimizer as it performs better than the original model (see the appendix).
In addition, the upper bound on the number of stations is computed in the same way as the MIP model instead of using a heuristic.
DP is better than MIP and CP, especially in large instances.

\subsubsection{Bin Packing}
We use instances in BPPLIB \cite{Delorme2018} and the MIP model \cite{Martello1990} extended with inequalities ensuring that bins are used in order of index and item $j$ is packed in the $j$-th bin or before as described in \citeauthor{Delorme2016}~\shortcite{Delorme2016}.
We implement a CP model using $\mathsf{Pack}$ while ensuring that item $j$ is packed in bin $j$ or before.
For MIP and CP models, the upper bound on the number of bins is computed by the first-fit decreasing heuristic.
We show the CP model in the appendix.
CP solves more instances than MIP and DP except for Scholl 1.
Similar to CVRP, without the precedence constraints of SALBP-1, CAASDy suffers from a large branching factor and quickly runs out of memory.

\subsubsection{MOSP}
We use instances in Constraint Modelling Challenge \cite{ConstraintModellingChallenge}, SCOOP Project, \citeauthor{Faggioli1998}~\shortcite{Faggioli1998}, and \citeauthor{Chu2009}~\shortcite{Chu2009}.
The MIP and CP models are proposed by \citeauthor{Martin2021}~\shortcite{Martin2021}.
From their two MIP models, we select MOSP-ILP-I as it solves more instances optimally in their paper.
DP solves more instances than MIP and CP in the Chu and Stuckey problem set, which results in higher coverage in total.

\subsubsection{Graph-Clear}
We generate instances using planar and random graphs in the same way as \citeauthor{Morin2018}~\shortcite{Morin2018}, where the number of nodes in a graph is 20, 30, or 40.
We use MIP and CP models proposed by \citeauthor{Morin2018}~\shortcite{Morin2018}.
From the two proposed CP models, we select CPN as it performs better in our setting.
DP solves more instances than MIP and CP.
While MIP and CP only solve instances with 20 nodes, DP solves all planar instances with 20 and 30 nodes, all random instances with 20 nodes, 5 out of 20 planar instances with 40 nodes, and 6 out of 25 random instances with 30 nodes.

\section{Discussion}
First, we compare DIDP with existing approaches to clarify its novelty.
Then, we summarize the significance of DIDP.

\subsection{Model-Based DP}
Little work has considered DP as a domain-independent model-based approach.
DP2PN2Solver \cite{Lew2006} is a C++/Java style modeling language with an associated DP solver that explicitly enumerates all reachable states.
Algebraic dynamic programming (ADP) \cite{Giegerich2002} is a framework to formulate a DP model using context-free grammar that was originally designed for bioinformatics and limited to problems on strings.
Although ADP has been extended to describe diverse DP models \cite{zuSiederdissen2015}, it is focused on bioinformatics applications.

\subsection{AI Planning}
Except for MOSP and graph-clear, the DP models above can be formulated as numeric planning problems, but resource variables, forced transitions, and a dual bound cannot be modeled.
We evaluated the PDDL models with numeric planners using A* with admissible heuristics \cite{Kuroiwa2022}, but they did not show competitive performance as they are not designed for such problems.
For MOSP, a classical planning model based on a different formulation was used in the International Planning Competition.\footnote{\url{https://ipc06.icaps-conference.org/deterministic/}}
A state-of-the-art planner, SymBA* \cite{Torralba2016} outperforms MIP but not CP or DP.
Picat \cite{Picat} is a logic-based programming language providing a DP solver and an associated AI planning module.
However, the solution method is restricted to backtracking, and it cannot model resource variables.

\subsection{Decision Diagram Solver}
The existing work most similar to ours is ddo, a decision diagrams (DD) solver that uses DP as a modeling interface \cite{Gillard2020}.
Ddo is not a generic DP solver as it requires a problem-specific merge operator for DD nodes in addition to a DP model. The merge operator is necessary for relaxed DDs and, consequently, dual bounds.
Defining a dual bound is optional in DyPDL and is done in the language of the model, not of the solver (i.e., a merge operator is only relevant to a DD-based solver).
Developing a domain-independent merge operator for ddo is an interesting direction for future work.
Since ddo was previously used in TSPTW, we evaluate it in our setting. While CAASDy solves 257 instances, ddo solves 179 instances (see the appendix for details). We do not evaluate ddo in other problems as the merge operators are not defined by previous work.

Hadook \cite{Gentzel2020} is a modeling language for decision diagrams developed for constraint propagation in CP.
Hadook is similar to DyPDL in that its formalism is based on a state transition system.

\subsection{Significance and Impact}
In summary, DIDP is novel in the following points:
it considers DP as a model-based approach, separating modeling and solving, for combinatorial optimization;
its modeling formalism, DyPDL, is explicitly designed to allow a user to incorporate implications of the problem definition, i.e., resource variables, forced transitions, state constraints, and a dual bound, in a DP model, following the standard in OR.
Our prototype solver shows state-the-of-art performance in multiple problem classes, as shown in the experimental result, which supports the significance of DIDP.
DIDP also bridges the gap between AI and OR communities:
DIDP enables researchers in AI planning and heuristic search to apply their methods to OR problems.

\section{Conclusion}
We proposed Domain-Independent Dynamic Programming (DIDP), a new model-based paradigm for combinatorial optimization.
We developed Dynamic Programming Description Language (DyPDL), a modeling language for DP, and Cost-Algebraic A* Solver for DyPDL (CAASDy), a prototype DyPDL solver.
Our solver outperforms MIP and CP in multiple combinatorial optimization problems.

While we formulated diverse DP models with DyPDL, there is significant room for extensions.
For example, dominance relationships and symmetry breaking based on other criteria may be desired for efficient DP models.
Also, there is a significant opportunity to improve our solver using state space search methods that have been developed in AI planning and heuristic search over the past two decades.
For example, while CAASDy uses a dual bound defined in a DP model as a heuristic function, most AI planners automatically compute heuristic functions.
Adapting AI planning methods to obtain a dual bound is one of our future plans.

\section{Acknowledgments}
This work was partially supported by the Natural Sciences and Engineering Research Council of Canada.

\section{Appendix: CP Models}
We present the new CP models used in the experimental evaluation.

\subsection{TSPTW}
We adapt a CP model for a single machine scheduling problem with time windows \cite{Booth2016} to TSPTW.
Let $x_i$ be an interval variable in a range $[a_i, b_i]$ with the length of $0$, representing visiting customer $i$.
\begin{align*}
    \min         & \sum_{i \in N} c_{i, \mathsf{Next}(x_i)}                                                 \\
    \text{s.t. } & \mathsf{NoOverlap}([x_0, ..., x_{n-1}], \{ c_{ij} \mid i, j \in N \}) &                  \\
                 & \mathsf{First}(x_0)                                                   &                  \\
                 & x_i: \mathsf{intervalVar}(0, [a_i, b_i])                              & \forall i \in N.
\end{align*}
The first constraint ensures that interval variables are ordered in a sequence, and for two consecutive variables $x_i$ and $x_j$, the start of $x_j$ must be at least $c_{ij}$ greater than the end of $x_i$.
In the objective, $\mathsf{Next}(x_i)$ is the interval variable next to $x_i$ in the sequence.
For the last variable, we let $\mathsf{Next}(x_i) = x_0$.
The second constraint ensures that the depot is visited first.

\subsection{SALBP-1}
For SALBP-1, we implement the CP model proposed by \citeauthor{Bukchin2018}~\shortcite{Bukchin2018} with the addition of the $\mathsf{Pack}$ global constraint \cite{Shaw2004}.
For an upper bound on the number of stations, instead of using a heuristic to compute it, we use $\bar{m} = \min \{ n, 2 \lceil \sum_{i \in N} t_i / c \rceil \}$ following the MIP model \cite{Ritt2018}.
Let $M = \{ 0, ..., \bar{m} - 1 \}$ be the set of stations.
Let $m$ be a decision variable representing the number of stations, $x_i$ be a decision variable representing the index of the station of task $i$, and $y_j$ be the sum of the processing times of tasks scheduled in station $j$.
The set of all direct and indirect predecessors of task $i$ is
\[
    \tilde{P}_i = \{ j \in N \mid j \in P_i \lor \exists k \in \tilde{P}_j, j \in \tilde{P}_k \}.
\]
The set of all direct and indirect successors of task $i$ is
\[
    \tilde{S}_i = \{ j \in N \mid i \in P_j \lor \exists k \in \tilde{S}_i, j \in \tilde{S}_k \}.
\]
Thus,
\[
    e_i = \left\lceil \frac{t_i + \sum_{k \in \tilde{P}_i} t_k}{c} \right\rceil
\]
is a lower bound on the number of stations required to schedule task $i$,
\[
    l_i = \left\lfloor \frac{t_i - 1 + \sum_{k \in \tilde{S}_i} t_k}{c} \right\rfloor
\]
is a lower bound on the number of stations between the station of task $i$ and the last station, and
\[
    d_{ij} = \left\lfloor \frac{t_i + t_j - 1 + \sum_{k \in \tilde{S}_i \cap \tilde{P}_j} t_k}{c} \right\rfloor
\]
is a lower bound on the number of stations between the stations of tasks $i$ and $j$.
\begin{align*}
    \min \quad   & m                                                                                                                            \\
    \text{s.t. } & \mathsf{Pack}(\{ y_j \mid j \in M \}, \{ x_i \mid i \in N \}, \{ t_i \mid i \in N \})                                        \\
                 & 0 \leq y_j \leq c ~~~ \quad \quad \quad \quad \quad \quad \quad \quad \quad \quad \quad \quad \forall j \in M                \\
                 & e_i - 1 \leq x_i \leq m - 1 - l_i ~~~ \quad \quad \quad \quad \quad \quad \forall i \in N                                    \\
                 & x_i + d_{ij} \leq x_j \quad \quad \forall j \in N, \forall i \in \tilde{P}_j,                                                \\
                 & \quad \quad \quad \quad \quad \quad \quad \quad \not\exists k \in \tilde{S}_i \cap \tilde{P}_j : d_{ij} \leq d_{ik} + d_{kj} \\
                 & m \in \mathbb{Z}                                                                                                             \\
                 & y_j \in \mathbb{Z}  ~~ \quad \quad \quad \quad \quad \quad \quad \quad \quad \quad \quad \quad \quad \quad \forall j \in M   \\
                 & x_i \in \mathbb{Z}  ~~ \quad \quad \quad \quad \quad \quad \quad \quad \quad \quad \quad \quad \quad \quad \forall i \in N.
\end{align*}
The first constraint ensures $x_i \in M$ and $\sum_{i \in N : x_i = j} t_i = y_j$.
The second constraint ensures that the sum of the processing times does not exceed the cycle time.
The third constraint states the lower and upper bounds on the index of the station of $i$.
The fourth constraint is an enhanced version of the precedence constraint using $d_{ij}$.

\subsection{Bin Packing}
For the CP model for bin packing, we also use $\mathsf{Pack}$.
In addition, we ensure that item $i$ is packed in the $i$-th or an earlier bin.
\begin{align*}
    \min         & \max_{i \in N} x_i + 1                                                                                                      \\
    \text{s.t. } & \mathsf{Pack}(\{ y_j \mid j \in M \}, \{ x_i \mid i \in N \}, \{ t_i \mid i \in N \})                                       \\
                 & 0 \leq y_j \leq c ~~~ \quad \quad \quad \quad \quad \quad \quad \quad \quad \quad \quad \quad \forall j \in M               \\
                 & 0 \leq x_i \leq i \quad \quad \quad \quad \quad \quad \quad \quad \quad \quad \quad \quad \quad \forall i \in N             \\
                 & y_j \in \mathbb{Z}  ~~ \quad \quad \quad \quad \quad \quad \quad \quad \quad \quad \quad \quad \quad \quad \forall j \in M  \\
                 & x_i \in \mathbb{Z}  ~~ \quad \quad \quad \quad \quad \quad \quad \quad \quad \quad \quad \quad \quad \quad \forall i \in N.
\end{align*}
We compute the upper bound $\bar{m}$ using the first-fit decreasing heuristic.

\section{Appendix: DP Models}
We prove the properties of the DP models assumed in the paper and provide the DyPDL representations of the models.

\subsection{Lower Bounds for SALBP-1 and Bin Packing}
We show that the lower bounds used in the DP models for SALBP-1 and bin packing are valid.
These lower bounds, LB1, LB2, and LB3 were originally proposed by \citeauthor{Scholl1997}~\shortcite{Scholl1997}.
The first lower bound, LB1, is originally defined as $\left\lceil \sum_{i \in N} t_i / c \right\rceil$.
This bound relaxes the problem by allowing to split a task across multiple stations.
In a state $\langle U, r \rangle$ ($\langle U, r, k \rangle$ for bin packing), we only need to schedule tasks in $U$, and we can schedule tasks in the current station, which has the remaining time of $r$.
Therefore, we use $\lceil (\sum_{i \in U} t _i - r) / c \rceil$ as a lower bound.

The second lower bound, LB2, is originally defined as $\sum_{i \in N} w^2_i + \lceil \sum_{i \in N} w'^2_i \rceil$, where $w^2_i = 1$ if $t_i > c / 2$ and $w'^2_i = 1 / 2$ if $t_i = c / 2$.
This bound only considers tasks $i$ with $t_i \geq 2 / c$.
The first term, the number of tasks $i$ with $t_i > 2 / c$, is a lower bound because other tasks cannot be scheduled in the same station as $i$.
For the remaining tasks, which have $t_i = 2 / c$, two tasks can be scheduled in the same station, which results in the second term.
In our model, in a state $\langle U, r \rangle$, we use the bound $\sum_{i \in U} w^2_i + \lceil \sum_{i \in U} w'^2_i \rceil$ if $r < c/2$ because we cannot schedule any tasks with $t_i \geq 2 / c$ in the current station.
If $r \geq c/2$, since we may use the current station, we subtract $1$ from the bound.

The second lower bound, LB3, is based on a similar idea to LB2.
It is originally defined as $\lceil \sum_{i \in N} w'^3_i \rceil$ and only considers tasks $i$ with $t_i \geq 3 / c$.
Therefore, we use the bound $\lceil \sum_{i \in U} w^3_i \rceil$ if $r < c/3$ and subtract $1$ from it otherwise.

\subsection{Cost-Algebra for MOSP and Graph-Clear}
We show that the cost expressions in the DP models for MOSP and graph-clear satisfy the property of cost-algebra.
In these models, the cost expressions are in the form of $\max\{ e_\tau(S), V(S[\![\tau]\!]) \}$ instead of $e_\tau(S) + V(S[\![\tau]\!])$.

A cost-algebra \cite{Edelkamp2005} is defined as a $6$-tuple $\langle A, \sqcup, \times, \preceq, \mathbf{0}, \mathbf{1} \rangle$ where $A$ is a set, $\times: A \times A \rightarrow A$ is a binary operator, $\preceq ~ \in A \times A$ is a binary relation, $\mathbf{0}, \mathbf{1} \in A$, and $\sqcup: 2^A \rightarrow A$ is an operator to select one element from a subset of $A$.
It must satisfy the following conditions.
\begin{enumerate}
    \item $\forall a, b \in A, a \times b \in A$
    \item $\forall a, b, c \in A, a \times (b \times c) = (a \times b) \times c$
    \item $\forall a \in A, a \times \mathbf{1} = \mathbf{1} \times a = a$
    \item $\forall a \in A, a \preceq a$
    \item $\forall a, b \in A, a \preceq b  \land b \preceq a \Rightarrow a = b$
    \item $\forall a, b, c \in A, a \preceq b \land b \preceq c \Rightarrow a \preceq c$
    \item $\forall a, b \in A, a \preceq b \lor b \preceq a$
    \item $\forall B \subseteq A, \forall b \in B, \sqcup B \preceq b$
    \item $\forall a \in A, a \preceq \mathbf{0}$ and $\mathbf{1} \preceq a$
    \item $\forall a, b, c \in A, a \preceq b \Rightarrow a \times c \preceq b \times c$ and $c \times a \preceq c \times b$
\end{enumerate}
Conditions 1-3 ensure that $\langle A, \times, \mathbf{1} \rangle$ is a monoid.
Conditions 4-7 ensure that $\preceq$ is a total order.
Condition 10 is called isotonicity.

\citeauthor{Edelkamp2005}~\shortcite{Edelkamp2005} proved that $\langle \mathbb{R}^+ \cup \{ +\infty \}, \min, +, \leq, +\infty, 0 \rangle$, which corresponds to the shortest path problem, satisfies the conditions.
We show that a tuple $\langle \mathbb{R}^+ \cup \{ +\infty \}, \min, \max, \leq, +\infty, 0 \rangle$ satisfies the conditions, which corresponds to the cost expresssions in the DP models for MOSP and graph-clear.
The tuple $\langle \mathbb{R}^+ \cup \{ +\infty \}, \max, 0 \rangle$ is a monoid since $\max \{ a, b \} \in \mathbb{R}$, $\max\{ a, \max\{ b, c \} \} = \max\{ \max\{ a, b \}, c \}$, and $\max \{ a, 0 \} = \max \{ 0, a \} = a$.
Conditions 4-9 hold since $\langle \mathbb{R}^+ \cup \{ +\infty \}, \min, +, \leq, +\infty, 0 \rangle$ is a cost-algebra.
For the isotonicity, $\max \{ a, c \} \leq \max \{ b, c \}$ and $\max \{ c, a \} \leq \max \{ c, b \}$ for $a \leq b$.

\subsection{DyPDL Representations}
We present DyPDL representations and YAML-DyPDL domain files of the DP models for CVRP in Table~\ref{tab:dypdl-cvryp} and Listing~\ref{lst:domain-cvrp}, for SALBP-1 in Table~\ref{tab:dypdl-salbp-1} and Listing~\ref{lst:domain-salbp-1}, for bin packing in Table~\ref{tab:dypdl-bin-packing} and Listing~\ref{lst:domain-bin-packing}, for MOSP in Table~\ref{tab:dypdl-mosp} and Listing~\ref{lst:domain-mosp}, and for graph-clear in Table~\ref{tab:dypdl-graph-clear} and Listing~\ref{lst:domain-graph-clear}.

\begin{table}
    \small
    \tabcolsep=0.3em
    \begin{tabular}{l|lll}
        $\mathcal{V}$                            & Type                                                                               & Objects                      & Preference       \\
        \hline
        $U$                                      & set                                                                                & customers $N$                &                  \\
        $i$                                      & element                                                                            & customers $N$                &                  \\
        $l$                                      & numeric                                                                            &                              & less             \\
        $k$                                      & numeric                                                                            &                              & less             \\
        \hline
        \hline
        $\mathcal{K}$                            & Type                                                                               & Indices                                         \\
        \hline
        $q$                                      & numeric                                                                            &                                                 \\
        $m$                                      & numeric                                                                            &                                                 \\
        $d_j$                                    & numeric                                                                            & $j \in N$                                       \\
        $c_{jp}$                                 & numeric                                                                            & $j, p \in N$                                    \\
        $c'_{jp}$                                & numeric                                                                            & $j, p \in N$                                    \\
        \hline
        \hline
        $S^0$                                    & \multicolumn{3}{l}{$\langle U = N \setminus \{ 0 \}, i = 0, l = 0, k = 1 \rangle$}                                                   \\
        $\mathcal{B}$                            & \multicolumn{3}{l}{$\{ \{ U = \emptyset, i = 0 \} \}$}                                                                               \\
        $\mathcal{C}$                            & \multicolumn{3}{l}{$\emptyset$}                                                                                                      \\
        $h$                                      & \multicolumn{3}{l}{$0$}                                                                                                              \\
        \hline
        \hline
        $\mathcal{T}$                            & $\mathsf{eff}$                                                                     & $\mathsf{cost}$              & $\mathsf{pre}$   \\
        \hline
        \multirow{3}{*}{visit $j$}               & $U \leftarrow U \setminus \{ j \}$                                                 & $c_{ij} + V(S[\![\tau]\!])$  & $j \in U$        \\
                                                 & $i \leftarrow j$                                                                   &                              & $l + d_j \leq q$ \\ & $l \leftarrow l + d_j$ &                  & \\
        \hline
        \multirow{4}{*}{visit $j$ via the depot} & $U \leftarrow U \setminus \{ j \}$                                                 & $c'_{ij} + V(S[\![\tau]\!])$ & $j \in U$        \\
                                                 & $i \leftarrow j$                                                                   &                              & $k < m$          \\ & $l \leftarrow d_j$ &                  & \\
                                                 & $k \leftarrow k + 1$                                                                                                                 \\
        \hline
        \multirow{2}{*}{return}                  & $i \leftarrow 0$                                                                   & $c_{i0} + V(S[\![\tau]\!])$  & $U = \emptyset$  \\
                                                 &                                                                                    &                              & $i \neq 0$
    \end{tabular}
    \caption{DyPDL representation of the DP model for CVRP. No forced transition exists in this model.}
    \label{tab:dypdl-cvryp}
\end{table}

In SALBP-1, \texttt{open-station} is a forced transition in the YAML-DyPDL domain file.
While it is not necessarily in theory because the other transitions are not applicable when \texttt{open-station} is applicable, it can be beneficial since a solver does not need to evaluate the preconditions of the other transitions.

\begin{table}
    \small
    \tabcolsep=0.6em
    \begin{tabular}{l|lll}
        $\mathcal{V}$               & Type                                                                                                                                                                                                                                & Objects                & Preference               \\
        \hline
        $U$                         & set                                                                                                                                                                                                                                 & tasks $N$              &                          \\
        $r$                         & numeric                                                                                                                                                                                                                             &                        & more                     \\
        \hline
        \hline
        $\mathcal{K}$               & Type                                                                                                                                                                                                                                & Objects                & Indices                  \\
        \hline
        $c$                         & numeric                                                                                                                                                                                                                             &                        &                          \\
        $t_i$                       & numeric                                                                                                                                                                                                                             &                        & $i \in N$                \\
        $P_i$                       & set                                                                                                                                                                                                                                 & tasks $N$              & $i \in N$                \\
        $w^2_i$                     & numeric                                                                                                                                                                                                                             &                        & $i \in N$                \\
        $w'^2_i$                    & numeric                                                                                                                                                                                                                             &                        & $i \in N$                \\
        $w^3_i$                     & numeric                                                                                                                                                                                                                             &                        & $i \in N$                \\
        \hline
        \hline
        $S^0$                       & \multicolumn{3}{l}{$\langle U = N, r = 0 \rangle$}                                                                                                                                                                                                                                      \\
        $\mathcal{B}$               & \multicolumn{3}{l}{$\{ \{ U = \emptyset \} \}$}                                                                                                                                                                                                                                         \\
        $\mathcal{C}$               & \multicolumn{3}{l}{$\emptyset$}                                                                                                                                                                                                                                                         \\
        $h$                         & \multicolumn{3}{l}{$\max \left\{ \begin{array}{l} \lceil (\sum_{i \in U} t_i - r) / c \rceil  \\ \sum_{u \in U} w^2_i + \lceil \sum_{i \in U} w'^2_i \rceil - l^2 \\ \lceil \sum_{i \in U} w^3_i \rceil- l^3  \end{array} \right.$}                                                     \\
        \hline
        \hline
        $\mathcal{T}$               & $\mathsf{eff}$                                                                                                                                                                                                                      & $\mathsf{cost}$        & $\mathsf{pre}$           \\
        \hline
        \multirow{3}{*}{assign $i$} & $U \leftarrow U \setminus \{ i \}$                                                                                                                                                                                                  & $V(S[\![\tau]\!])$     & $i \in U$                \\
                                    & $r \leftarrow r - t_i$                                                                                                                                                                                                              &                        & $P_i \cap U = \emptyset$ \\
                                    &                                                                                                                                                                                                                                     &                        & $r \geq t_i$             \\
        \hline
        open a station              & $r \leftarrow c$                                                                                                                                                                                                                    & $1 + V(S[\![\tau]\!])$ & $U' = \emptyset$         \\
    \end{tabular}
    \caption{DyPDL representation of the DP model for SALBP-1, where $U' = \{ i \in U \mid P_i \cap U = \emptyset \land r \geq t_i \}$, $l^2 = 1$ if $r \geq c/2$ and $l^2 = 0$ otherwise, and $l^3 = 1$ if $r \geq c/3$ and $l^3=0$ otherwise.}
    \label{tab:dypdl-salbp-1}
\end{table}

\begin{table}
    \small
    \tabcolsep=0.4em
    \begin{tabular}{l|llll}
        $\mathcal{V}$                  & Type                                                                                                                                                                                                                                & Objects                & Preference                            \\
        \hline
        $U$                            & set                                                                                                                                                                                                                                 & items $N$              &                                       \\
        $r$                            & numeric                                                                                                                                                                                                                             &                        & more                                  \\
        $k$                            & element                                                                                                                                                                                                                             & items $N$              & less                                  \\
        \hline
        \hline
        $\mathcal{K}$                  & Type                                                                                                                                                                                                                                & Indices                                                        \\
        \hline
        $c$                            & numeric                                                                                                                                                                                                                             &                                                                \\
        $t_i$                          & numeric                                                                                                                                                                                                                             & $i \in N$                                                      \\
        $w^2_i$                        & numeric                                                                                                                                                                                                                             & $i \in N$                                                      \\
        $w'^2_i$                       & numeric                                                                                                                                                                                                                             & $i \in N$                                                      \\
        $w^3_i$                        & numeric                                                                                                                                                                                                                             & $i \in N$                                                      \\
        \hline
        \hline
        $S^0$                          & \multicolumn{4}{l}{$\langle U = N, r = 0, k = 0 \rangle$}                                                                                                                                                                                                                                            \\
        $\mathcal{B}$                  & \multicolumn{4}{l}{$\{ \{ U = \emptyset \} \}$}                                                                                                                                                                                                                                                      \\
        $\mathcal{C}$                  & \multicolumn{4}{l}{$\emptyset$}                                                                                                                                                                                                                                                                      \\
        $h$                            & \multicolumn{4}{l}{$\max \left\{ \begin{array}{l} \lceil (\sum_{i \in U} t_i - r) / c \rceil  \\ \sum_{u \in U} w^2_i + \lceil \sum_{i \in U} w'^2_i \rceil - l^2 \\ \lceil \sum_{i \in U} w^3_i \rceil- l^3  \end{array} \right.$}                                                                  \\
        \hline
        \hline
        $\mathcal{T}$                  & $\mathsf{eff}$                                                                                                                                                                                                                      & $\mathsf{cost}$        & $\mathsf{pre}$    & $\mathsf{forced}$ \\
        \hline
        \multirow{3}{*}{pack $i$}      & $U \leftarrow U \setminus \{ i \}$                                                                                                                                                                                                  & $V(S[\![\tau]\!])$     & $i \in U$         & $\bot$            \\
                                       & $r \leftarrow r - t_i$                                                                                                                                                                                                              &                        & $r \geq t_i$                          \\
                                       &                                                                                                                                                                                                                                     &                        & $i + 1 \geq k$                        \\
        \hline
        \multirow{3}{*}{open with $i$} & $U \leftarrow U \setminus \{ i \}$                                                                                                                                                                                                  & $1 + V(S[\![\tau]\!])$ & $U^1 = \emptyset$ & $\top$            \\
                                       & $r \leftarrow c - t_i$                                                                                                                                                                                                              &                        & $i \in U$                             \\
                                       & $k \leftarrow k + 1$                                                                                                                                                                                                                &                        & $i \geq k$                            \\
    \end{tabular}
    \caption{DyPDL representation of the DP model for bin packing, where $U^1 = \{ i \in U \mid r \geq t_i \land i + 1 \geq k \}$, $l^2 = 1$ if $r \geq c/2$ and $l^2 = 0$ otherwise, and $l^3 = 1$ if $r \geq c/3$ and $l^3=0$ otherwise.}
    \label{tab:dypdl-bin-packing}
\end{table}

\begin{table}
    \small
    \tabcolsep=0.25em
    \begin{tabular}{l|lll}
        $\mathcal{V}$              & Type                                                       & Objects                                                                                                     &                \\
        \hline
        $R$                        & set                                                        & customers $C$                                                                                               &                \\
        $O$                        & set                                                        & customers $C$                                                                                               &                \\
        \hline
        \hline
        $\mathcal{K}$              & Type                                                       & Objects                                                                                                     & Indices        \\
        \hline
        $N_c$                      & set                                                        & customers $C$                                                                                               & $c \in C$      \\
        \hline
        \hline
        $S^0$                      & \multicolumn{3}{l}{$\langle R = C, O = \emptyset \rangle$}                                                                                                                                \\
        $\mathcal{B}$              & \multicolumn{3}{l}{$\{ \{ R = \emptyset \} \}$}                                                                                                                                           \\
        $\mathcal{C}$              & \multicolumn{3}{l}{$\emptyset$}                                                                                                                                                           \\
        $h$                        & \multicolumn{3}{l}{$0$}                                                                                                                                                                   \\
        \hline
        \hline
        $\mathcal{T}$              & $\mathsf{eff}$                                             & $\mathsf{cost}$                                                                                             & $\mathsf{pre}$ \\
        \hline
        \multirow{2}{*}{close $c$} & $R \leftarrow R \setminus \{ c \}$                         & $\max \left\{ \begin{array}{l} V(S[\![\tau]\!]) \\ |(O \cap R) \cup (N_c \setminus O)| \end{array} \right.$ & $c \in R$      \\
                                   & $O \leftarrow O \cup N_c$                                  &                                                                                                             &                \\
    \end{tabular}
    \caption{DyPDL representation of the DP model for MOSP. No forced transition exists in this model.}
    \label{tab:dypdl-mosp}
\end{table}

\begin{table}
    \small
    \tabcolsep=0.5em
    \begin{tabular}{l|lll}
        $\mathcal{V}$ & Type                                                & Objects                               &                \\
        \hline
        $C$           & set                                                 & nodes $N$                             &                \\
        \hline
        \hline
        $\mathcal{K}$ & Type                                                & Objects                               & Indices        \\
        \hline
        $N$           & set                                                 & nodes $N$                             &                \\
        $a_c$         & numeric                                             &                                       & $c \in N$      \\
        $b_{ij}$      & numeric                                             &                                       & $i, j \in N$   \\
        \hline
        \hline
        $S^0$         & \multicolumn{3}{l}{$\langle C = \emptyset \rangle$}                                                          \\
        $\mathcal{B}$ & \multicolumn{3}{l}{$\{ \{ C = N \} \}$}                                                                      \\
        $\mathcal{C}$ & \multicolumn{3}{l}{$\emptyset$}                                                                              \\
        $h$           & \multicolumn{3}{l}{$0$}                                                                                      \\
        \hline
        \hline
        $\mathcal{T}$ & $\mathsf{eff}$                                      & $\mathsf{cost}$                       & $\mathsf{pre}$ \\
        \hline
        sweep $c$     & $C \leftarrow C \cup \{ c \}$                       & $\max\{ V(S[\![\tau]\!]), e(c, S) \}$ & $c \notin C$   \\
    \end{tabular}
    \caption{DyPDL representation of the DP model for graph-clear, where $e(c, S) = a_c + \sum_{i \in N} b_{ci} + \sum_{i \in C} \sum_{j \in \overline{C} \setminus \{ c \}} b_{ij}$. No forced transition exists in this model. }
    \label{tab:dypdl-graph-clear}
\end{table}

\begin{listing}[tb]
\begin{lstlisting}
objects:
  - customer
state_variables:
  - name: U
    type: set
    object: customer
  - name: i
    type: element
    object: customer
  - name: l
    type: integer
    preference: less
  - name: k
    type: integer
    preference: less
tables:
  - name: q
    type: integer
  - name: m
    type: integer
  - name: d
    type: integer
    args: [customer]
  - name: c
    type: integer
    args: [customer, customer]
    default: 0
  - name: c-via-depot
    type: integer
    args: [customer, customer]
base_cases:
  - [(is_empty U), (= i 0)]
reduce: min
cost_type: integer
transitions:
  - name: visit
    parameters: [{ name: j, object: U }]
    preconditions: [(<= (+ l (d j)) q)]
    effect:
      U: (remove j U)
      i: j
      l: (+ l (d j))
    cost: (+ cost (c i j))
  - name: visit-via-depot
    parameters: [{ name: j, object: U }]
    preconditions: [(< k m)]
    effect:
      U: (remove j U)
      i: j
      l: (d j)
      k: (+ k 1)
    cost: (+ cost (+ c-via-depot i j))
  - name: return
    preconditions:
      - (is_empty U)
      - (!= i 0)
    effect:
      i: 0
    cost: (+ cost (c i 0))
dual_bounds:
  - 0
\end{lstlisting}
    \caption{YAML-DyPDL domain file for CVRP.}
    \label{lst:domain-cvrp}
\end{listing}

\begin{listing}[!htb]
\begin{lstlisting}
objects:
  - task
state_variables:
  - name: U
    type: set
    object: task
  - name: r
    type: integer
    preference: greater
tables:
  - { name: c, type: integer }
  - name: t
    type: integer
    args: [task]
  - name: P
    type: set
    object: task
    args: [task]
  - name: w2_1
    type: integer
    args: [task]
  - name: w2_2
    type: continuous
    args: [task]
  - name: w3
    type: continuous
    args: [task]
base_cases:
  - - (is_empty U)
reduce: min
cost_type: integer
transitions:
  - name: assign
    parameters: [{ name: i, object: U }]
    preconditions:
      - (is_empty
          (intersection U (P i)))
      - (<= (t i) r)
    effect:
      U: (remove i U)
      r: (- r (t i))
    cost: cost
  - name: open-station
    forced: true
    preconditions:
      - forall: [{ name: i, object: U }]
        condition: >
          (or
            (> (t i) r)
            (> |(intersection U (P i))|
               0))
    effect:
      r: c
    cost: (+ cost 1)
dual_bounds:
  - (ceil (/ (- (sum t U) r) c))
  - (- (+ (sum w2_1 U)
          (ceil (sum w2_2 U)))
       (if (>= r (/ c 2.0)) 1 0))
  - (- (ceil (sum w3 U))
       (if (>= r (/ c 3.0)) 1 0))
\end{lstlisting}
    \caption{YAML-DyPDL domain file for SALBP-1.}
    \label{lst:domain-salbp-1}
\end{listing}

\begin{listing}[!htb]
\begin{lstlisting}
objects:
  - item
state_variables:
  - name: U
    type: set
    object: item
  - name: r
    type: integer
    preference: greater
  - name: k
    type: element
    object: item
    preference: less
tables:
  - name: c
    type: integer
  - name: t
    type: integer
    args: [item]
  - name: w2_1
    type: integer
    args: [item]
  - name: w2_2
    type: continuous
    args: [item]
  - name: w3
    type: continuous
    args: [item]
base_cases:
  - - (is_empty U)
reduce: min
cost_type: integer
transitions:
  - name: pack
    parameters: [{name: i, object: U}]
    preconditions:
      - (<= (t i) r)
      - (>= (+ i 1) k)
    effect:
      U: (remove i U)
      r: (- r (t i))
    cost: cost
  - name: open-with
    forced: true
    parameters: [{name: i, object: U}]
    preconditions:
      - (>= i k)
      - forall: [{name: j, object: U}]
        condition: (> (t j) r)
    effect:
      U: (remove i U)
      r: (- c (t i))
      k: (+ 1 k)
    cost: (+ cost 1)
dual_bounds:
  - (ceil (/ (- (sum t U) r) c))
  - (- (+ (sum w2_1 U)
          (ceil (sum w2_2 U))
       (if (>= r (/ c 2.0)) 1 0))
  - (- (ceil (sum w3 U))
       (if (>= r (/ c 3.0)) 1 0))
\end{lstlisting}
    \caption{YAML-DyPDL domain file for bin packing.}
    \label{lst:domain-bin-packing}
\end{listing}

\begin{listing}[!htb]
\begin{lstlisting}
objects:
  - customer
state_variables:
  - name: R
    type: set
    object: customer
  - name: O
    type: set
    object: customer
tables:
  - name: "N"
    type: set
    object: customer
    args:
      - customer
base_cases:
  - - (is_empty R)
reduce: min
cost_type: integer
transitions:
  - name: close
    parameters:
      - name: c
        object: R
    effect:
      R: (remove c R)
      O: (union O (N c))
    cost: >
      (max cost
        |(union (intersection O R)
                (difference (N c) O))|)
dual_bounds:
  - 0
\end{lstlisting}
    \caption{YAML-DyPDL domain file for MOSP.}
    \label{lst:domain-mosp}
\end{listing}

\begin{listing}[!htb]
\begin{lstlisting}
objects:
  - node
state_variables:
  - name: C
    type: set
    object: node
tables:
  - name: "N"
    type: set
    object: node
  - name: a
    type: integer
    args:
      - node
  - name: b
    type: integer
    args:
      - node
      - node
    default: 0
base_cases:
  - - (is_subset N C)
reduce: min
cost_type: integer
transitions:
  - name: sweep
    parameters:
      - name: c
        object: node
    preconditions:
      - (not (is_in c C))
    effect:
      C: (add c C)
    cost: >
      (max cost
           (+ (a c)
              (+ (sum b c N)
              (sum b C (remove c ~C)))))
dual_bounds:
  - 0
\end{lstlisting}
    \caption{YAML-DyPDL domain file for graph-clear.}
    \label{lst:domain-graph-clear}
\end{listing}

\section{Appendix: Experimental Results}
In addition to the number of instances solved to optimality, we also evaluate the computational time to find an optimal solution.
We take the average time over instances solved by all methods.
Furthermore, we evaluate the best lower bound found by the algorithm.
In CAASDy, the minimum $f$-value of states in the open list is a lower bound on the optimal solution.
For each instance, we compute the ratio of the lower bound found by a method to the best lower bound found by all the methods.
Concretely, if there are methods $1...,n$, and method $i$ finds a lower bound $L_i$, then, the lower bound ratio is defined as $\frac{L_i}{\max_{j=1,...,n} L_j}$.
Thus, higher is better and 1.0 is the maximum.

For TSPTW, we also evaluate the decision diagram-based solver, ddo \cite{Gillard2020} since it was previously used in TSPTW.
We use the `barrier` solver of ddo\footnote{\url{https://github.com/vcoppe/ddo-barrier}} \cite{Vianney2022}.
Since the original version is used to solve problems to minimize the makespan objective, we modified the code so that it minimizes the total travel time.

\begin{table*}[!htb]
    \small
    \centering
    \begin{tabular}{l|rrr|rrr|rrr|rrr}
                                        & \multicolumn{3}{c|}{MIP} & \multicolumn{3}{c|}{CP} & \multicolumn{3}{c|}{Ddo} & \multicolumn{3}{c}{DP}                                                                                                                      \\
    \hline
    TSPTW                               & \#                       & time                    & LB                       & \#                     & time          & LB            & \#         & time & LB            & \#            & time           & LB            \\
    \hline
    Dumas (135)                         & 121                      & 0.39                    & 0.97                     & 36                     & 52.61         & 0.43          & 114        & 0.11 & 0.86          & \textbf{135}  & \textbf{0.04}  & \textbf{1.00} \\
    GDE (130)                           & 71                       & 1.38                    & 0.71                     & 4                      & 289.20        & 0.16          & 35         & 1.53 & \textbf{0.73} & \textbf{77}   & \textbf{0.07}  & 0.71          \\
    OT (25)                             & \textbf{0}               & -                       & 0.00                     & \textbf{0}             & -             & 0.66          & \textbf{0} & -    & 0.32          & \textbf{0}    & -              & \textbf{0.67} \\
    AFG (50)                            & 33                       & 546.92                  & 0.80                     & 7                      & 16.02         & 0.40          & 30         & 0.30 & 0.81          & \textbf{45}   & \textbf{0.04}  & \textbf{0.95} \\
    \hline
    Total (340)                         & 225                      & 61.20                   & 0.77                     & 47                     & 69.58         & 0.34          & 179        & 0.26 & 0.77          & \textbf{257}  & \textbf{0.04}  & \textbf{0.86} \\
    \hline
    \hline
    CVRP                                & \#                       & time                    & LB                       & \#                     & time          & LB            & \#         & time & LB            & \#            & time           & LB            \\
    \hline
    A, B, E, F, P (90)                  & \textbf{26}              & -                       & \textbf{0.94}            & 0                      & -             & 0.05          & -          & -    & -             & 4             & -              & 0.22          \\
    \hline
    \hline
    SALBP-1                             & \#                       & time                    & LB                       & \#                     & time          & LB            & \#         & time & LB            & \#            & time           & LB            \\
    \hline
    Small (525)                         & \textbf{525}             & 0.25                    & \textbf{1.00}            & \textbf{525}           & 0.19          & \textbf{1.00} & -          & -    & -             & \textbf{525}  & \textbf{0.04}  & \textbf{1.00} \\
    Medium (525)                        & \textbf{518}             & 35.28                   & \textbf{1.00}            & 501                    & 15.30         & \textbf{1.00} & -          & -    & -             & 509           & \textbf{2.38}  & 0.99          \\

    Large (525)                         & 317                      & 103.44                  & 0.75                     & 404                    & 3.37          & \textbf{0.99} & -          & -    & -             & \textbf{414}  & \textbf{1.20}  & 0.88          \\
    Very large (525)                    & 0                        & -                       & 0.00                     & 155                    & -             & \textbf{0.97} & -          & -    & -             & \textbf{204}  & -              & 0.54          \\
    \hline
    Total (2100)                        & 1360                     & 37.31                   & 0.69                     & 1585                   & 6.47          & \textbf{0.99} & -          & -    & -             & \textbf{1652} & \textbf{1.17}  & 0.85          \\
    \hline
    \hline
    Bin Packing                         & \#                       & time                    & LB                       & \#                     & time          & LB            & \#         & time & LB            & \#            & time           & LB            \\
    \hline
    Falkenauer U (80)                   & 25                       & 64.49                   & 0.94                     & \textbf{36}            & 2.29          & \textbf{1.00} & -          & -    & -             & 33            & \textbf{1.61}  & 0.47          \\
    Falkenauer T (80)                   & 37                       & 147.01                  & \textbf{1.00}            & \textbf{56}            & 8.16          & \textbf{1.00} & -          & -    & -             & 27            & \textbf{7.19}  & 0.40          \\

    Scholl 1 (720)                      & \textbf{605}             & 16.88                   & 0.95                     & 533                    & 19.13         & \textbf{1.00} & -          & -    & -             & 517           & \textbf{15.70} & 0.85          \\

    Scholl 2 (480)                      & 354                      & 34.11                   & 0.97                     & \textbf{445}           & \textbf{0.37} & \textbf{1.00} & -          & -    & -             & 335           & 3.12           & 0.73          \\

    Scholl 3 (10)                       & 0                        & -                       & \textbf{1.00}            & \textbf{1}             & -             & \textbf{1.00} & -          & -    & -             & 0             & -              & 0.09          \\
    Falkenauer U                        & 25                       & 64.49                   & 0.94                     & \textbf{36}            & 2.29          & \textbf{1.00} & -          & -    & -             & 33            & \textbf{1.61}  & 0.47          \\
    W\"{a}scher (17)                    & 2                        & 788.96                  & \textbf{1.00}            & \textbf{10}            & 3.22          & \textbf{1.00} & -          & -    & -             & \textbf{10}   & \textbf{1.29}  & 0.62          \\
    Schwerin 1 (100)                    & 80                       & -                       & \textbf{1.00}            & \textbf{96}            & -             & \textbf{1.00} & -          & -    & -             & 0             & -              & 0.06          \\
    Schwerin 2 (100)                    & 54                       & -                       & \textbf{1.00}            & \textbf{61}            & -             & \textbf{1.00} & -          & -    & -             & 0             & -              & 0.09          \\
    Hard 28 (28)                        & \textbf{0}               & -                       & \textbf{1.00}            & \textbf{0}             & -             & \textbf{1.00} & -          & -    & -             & \textbf{0}    & -              & 0.32          \\
    \hline
    Total (1615)                        & 1157                     & 30.89                   & 0.97                     & \textbf{1238}          & 11.09         & \textbf{1.00} & -          & -    & -             & 922           & \textbf{10.19} & 0.66          \\

    \hline
    \hline
    MOSP                                & \#                       & time                    & LB                       & \#                     & time          & LB            & \#         & time & LB            & \#            & time           & LB            \\
    \hline
    Constraint Modelling Challenge (46) & 41                       & 10.32                   & 0.94                     & \textbf{44}            & 4.53          & 0.96          & -          & -    & -             & \textbf{44}   & \textbf{0.06}  & \textbf{1.00} \\
    SCOOP Project (24)                  & 8                        & 144.30                  & 0.64                     & \textbf{23}            & 0.50          & \textbf{0.99} & -          & -    & -             & 16            & \textbf{0.04}  & 0.95          \\

    Faggioli and Bentivoglio (300)      & 130                      & 88.66                   & 0.75                     & \textbf{300}           & 1.77          & \textbf{1.00} & -          & -    & -             & 298           & \textbf{0.04}  & \textbf{1.00} \\
    Chu and Stuckey (200)               & 44                       & 98.50                   & 0.47                     & 70                     & 58.10         & 0.49          & -          & -    & -             & \textbf{125}  & \textbf{0.07}  & \textbf{1.00} \\

    \hline
    Total (570)                         & 223                      & 76.94                   & 0.66                     & 437                    & 10.57         & 0.82          & -          & -    & -             & \textbf{483}  & \textbf{0.05}  & \textbf{1.00} \\

    \hline\hline
    Graph-Clear                         & \#                       & time                    & LB                       & \#                     & time          & LB            & \#         & time & LB            & \#            & time           & LB            \\
    \hline
    Planar (60)                         & 16                       & 447.49                  & 0.98                     & 1                      & 297.12        & 0.82          & -          & -    & -             & \textbf{20}   & \textbf{0.18}  & \textbf{1.00} \\
    Random (75)                         & 8                        & 65.35                   & 0.85                     & 3                      & 9.44          & 0.52          & -          & -    & -             & \textbf{25}   & \textbf{0.41}  & \textbf{1.00} \\
    \hline
    Total (135)                         & 24                       & 160.89                  & 0.91                     & 4                      & 81.36         & 0.65          & -          & -    & -             & \textbf{45}   & \textbf{0.36}  & \textbf{1.00} \\

    \hline
    \hline
    Average ratio                       & 0.48                     &                         &                          & 0.41                   &               &               & -          &      &               & \textbf{0.59}
\end{tabular}

    \caption{
        Number of instances solved to optimality (`\#'), the time to solve averaged over instances solved by all the methods (`time'), and the average lower bound ratio to the best lower bound found by all the methods (`LB').
        `Average ratio' is the ratio of optimally solved instances in each problem class averaged over all problem classes.
    }
\end{table*}

\clearpage

\fontsize{9.0pt}{10.0pt} \selectfont
\bibliography{reference-short}

\end{document}